\documentclass[runningheads]{llncs}

\usepackage{eccv}

\usepackage{eccvabbrv}

\usepackage{graphicx}
\usepackage{booktabs}

\usepackage[accsupp]{axessibility}  

\usepackage{hyperref}

\usepackage{orcidlink}

\usepackage{epsfig}
\usepackage{xspace}
\usepackage{blindtext}
\usepackage{caption}
\usepackage{amsfonts}
\usepackage[inline]{enumitem}
\usepackage{algorithmic}
\usepackage{textcomp}
\usepackage{multirow}
\usepackage{adjustbox}
\usepackage{etoolbox}
\usepackage{dsfont}
\usepackage{float}
\usepackage{lipsum}
\usepackage{stfloats}
\usepackage{multicol}
\usepackage{bm}
\usepackage{array}
\usepackage{tabulary}
\usepackage{paralist}

\usepackage{bbding}
\usepackage{pifont}
\usepackage{colortbl}
\usepackage{diagbox}
\usepackage{wrapfig}
\usepackage{caption}
\usepackage[normalem]{ulem}

\definecolor{tableHeadGray}{gray}{.9}

\usepackage[capitalize]{cleveref}
\crefname{section}{Sec.}{Secs.}
\Crefname{section}{Section}{Sections}
\Crefname{table}{Table}{Tables}
\crefname{table}{Tab.}{Tabs.}

\newcommand{\PAR}[1]{\noindent{\bf #1}}

\newcommand*{\shortname}{EvTemMap\@\xspace}
\newcommand*{\ourdataset}{TemMat dataset\@\xspace}
\newcommand*{\ourhardware}{AT-DVS\@\xspace}
\newcommand*{\timeimage}{Temporal Matrix\@\xspace}
\newcommand*{\ipe}{IPE\@\xspace}

\makeatletter
\newcommand{\printfnsymbol}[1]{%
  \textsuperscript{\@fnsymbol{#1}}%
}
\makeatother

\begin{document}
%
\title{Temporal-Mapping Photography for \\ Event Cameras}


\author{Yuhan Bao\inst{1}\thanks{Authors contributed equally to this work.}\orcidlink{0009-0000-3833-4565} \and
Lei Sun\inst{1,2}\printfnsymbol{1}\orcidlink{0000-0001-7310-5565} \and
Yuqin Ma\inst{1}\orcidlink{0009-0000-7235-6477} \and
Kaiwei Wang\inst{1}\orcidlink{}}


\authorrunning{Y.~Bao et al.}

\institute{National Research Center for Optical Instrumentation, Zhejiang University \and INSAIT, Sofia University \\
\email{wangkaiwei@zju.edu.cn, leo\_sun@zju.edu.cn}}

\maketitle

\begin{abstract}
Event cameras, or Dynamic Vision Sensors (DVS) are novel neuromorphic sensors that capture brightness changes as a continuous stream of ``events'' rather than traditional intensity frames. 
Converting sparse events to dense intensity frames faithfully has long been an ill-posed problem.
Previous methods have primarily focused on converting events to video in dynamic scenes or with a moving camera.
In this paper, for the first time, we realize events to dense intensity image conversion using a stationary event camera in static scenes with a transmittance adjustment device for brightness modulation.
Different from traditional methods that mainly rely on event integration, the proposed Event-Based Temporal Mapping Photography (\shortname) measures the time of event emitting for each pixel. Then, the resulting \timeimage is converted to an intensity frame with a temporal mapping neural network.
At the hardware level, the proposed \shortname is implemented by combining a transmittance adjustment device with a DVS, named Adjustable Transmittance Dynamic Vision Sensor (\ourhardware).
Additionally, we collected \ourdataset under various conditions including low-light and high dynamic range scenes.
The experimental results showcase the high dynamic range, fine-grained details, and high-grayscale resolution of the proposed \shortname.
The code and dataset are available in \url{https://github.com/YuHanBaozju/EvTemMap}.

\end{abstract}

\section{Introduction}
\label{sec:intro}
Event cameras offer high temporal resolution and dynamic range, swiftly responding to brightness changes through positive and negative events.
While the distinctive asynchronous data output from event cameras diminishes data redundancy, it concurrently discards the grayscale information of the image.
To capture grayscale images, a dynamic and active pixel vision sensor (DAVIS)~\cite{brandli2014240} integrates an active pixel sensor (APS) and a DVS sensor on a single chip. This integration allows the sensor to output both grayscale frames and events. The fusion of grayscale images with events facilitates various computer vision tasks.
Many event-based image/video enhancement tasks rely on such devices, such as motion deblur~\cite{sun2022event,zhang2022unifying,jiang2020learning,lin2020learning_event_video_deblur}, video interpolation~\cite{sun2023event,tulyakov2021time}, high dynamic range (HDR) imaging~\cite{messikommer2022multi}, auto-focusing~\cite{Bao:23} \etc. 
These tasks leverage the high temporal resolution and dynamic range of events to enhance grayscale images. 


However, due to the integration of multiple imaging sensors on a single chip, DAVISs have lower fill factors, and smaller pixel resolutions~\cite{gallego2020event}. Additionally, the grayscale images produced by the DAVIS device are consistent with traditional images and lack the high temporal resolution and dynamic range of DVS. 
Thus, to leverage the advantages of event cameras, some works aim to utilize events from pure DVS to reconstruct videos. Previous reconstruction approaches are based on hand-crafted priors and strong assumptions about the imaging process as well as the statistics of natural images~\cite{munda2018real}. 
Recent work, such as E2VID~\cite{rebecq2019events,stoffregen2020reducing}, trains a recurrent neural network to reconstruct videos from a stream of events. 
However, these methods suffer from grayscale distortion, lack of texture detail, and failure on static scenes. In contrast to E2VID, tailored for dynamic scenes, our objective is to capture diverse scenes, encompassing both static and dynamic ones, akin to traditional photography.


Since the temporal dimension of events exhibits theoretically infinite description levels, with 1 $\mu$s representing one level, can we utilize time to measure grayscale? 
Imagine a scene: the event camera is initially in complete darkness, and when the lighting gradually increases, the event camera captures positive events as each pixel gets brighter. Each pixel lights up at a different rate, depending on its corresponding point's light intensity in the scene.
As the event sensor initiates from darkness and gradually intensifies, the time each pixel takes to reach a specified threshold (i.e., the timestamp of the initial positive event, \ipe) indicates its brightening speed.
Hence, through the measurement of the timestamp associated with each \ipe, we can derive the light intensity of the corresponding point within the scene.

Building upon the mentioned basic concept, we introduce our \shortname. Initially, we block all incoming light to the event camera. Then, as we progressively raise the transmittance of the optical system, the brightness of each pixel increases at different rates. 
Throughout this process, we record the timestamp of \ipe for each pixel, forming a \timeimage. 
Subsequently, the \timeimage undergoes time-intensity mapping and finally produces a grayscale image. However, due to hardware constraints, the time-intensity mapping is not as ideal as we derive. Various factors such as uncertain contrast threshold, timestamp error, \etc degrade the final image. To address this issue, we design a temporal and spatial degradation space, especially for event-to-image conversion. Trained on the designed degradation model, our temporal-mapping network converts \timeimage to high-quality, high-grayscale-resolution, and high-dynamic-range image faithfully.
Additionally, we record \timeimage across diverse scenes and compile a dataset named \ourdataset. Through extensive comparisons with existing event-to-video methods, we demonstrate the superiority of the proposed \shortname.

In summary, we make the following main contributions:
\begin{compactitem}
    \item We revisit the mechanism of the event camera, and propose event-based temporal-mapping photography (\shortname)to convert events from a stationary \ourhardware in static scenes to an image, which is the first work in this area to the best of our knowledge.
    \item We further explore the temporal and spatial degradation, design a degradation model and train a temporal mapping network for our \shortname.
    \item We present \ourdataset consisting of diverse scenarios, varying illuminance levels, and high dynamic range scenes. Experiments on the dataset underscore the superiority of the proposed method.
\end{compactitem}

\section{Related Work}
\label{sec:related_work}

\subsection{Event to Video Reconstruction}
\label{sec:e2vid}
\PAR{Non-learning-based methods.}~The pioneering work~\cite{cook2011interacting} demonstrated the feasibility of extracting intensity information from event data and reconstructing images from events captured by an event camera traversing a stationary scene. Their approach mainly depends on the inherent connection between intensity gradients and optical flow, utilizing each event to formulate equations for brightness constancy.
Subsequently, event integration was proposed to be independent of assumptions regarding scene structure or motion dynamics~\cite{munda2018real} by Munda \etal. Scheerlinck \etal~\cite{scheerlinck2018continuous} defined the reconstruction as an energy minimization problem and applied a high-pass filter to events before integration. However, these methods frequently encounter "bleeding edges" caused by spatially variant contrast thresholds.

\PAR{Learning-based Methods.}~With the advancement of deep learning, Barua \etal. \cite{barua2016direct} utilized K-SVD~\cite{aharon2006k} on simulated data to train a dictionary mapping small patches of integrated events to image gradients. Subsequently, they employed Poisson integration for intensity image reconstruction. Rebecq \etal~\cite{rebecq2019high} designed an end-to-end recurrent model trained with synthetic events, resulting in the high-speed video reconstruction method E2VID. Subsequent works~\cite{scheerlinck2020fast,stoffregen2020reducing,cadena2021spade,zhang2020learning} maintained this paradigm while improving efficiency~\cite{scheerlinck2020fast}, fine-grained details~\cite{cadena2021spade}, and performance in low-light conditions~\cite{zhang2020learning}. More recently, Paredes-Vall{'e}s~\etal. \cite{paredes2021back} extended E2VID to self-supervised learning by introducing optical flow as a constraint. However, all these methods only recover intensity up to an unknown initial image, leading to ``ghosting'' effects and non-uniform grayscale distribution. Moreover, these methods are primarily applicable in dynamic scenes, where either the camera or the scene exhibits motion.

\subsection{High Dynamic Range Imaging}
\label{sec:HDR}
\PAR{Bracket exposure HDR.}~The most common method for achieving HDR photography is to use different exposure parameters to expose the same scene multiple times, obtain multiple low dynamic range (LDR) images, and merge them to reconstruct an HDR image~\cite{kalantari2017deep}. This technique, commonly referred to as bracket exposure, poses a crucial and challenging step in aligning multiple LDR images~\cite{wang2021deep}. Among the representative alignment methods are: Optical Flow-based Image Alignment~\cite{kalantari2017deep,yan2019multi,dosovitskiy2015flownet}, Direct Feature Concatenation~\cite{wu2018deep,chaudhari2021merging,niu2021hdr}, Correlation-Guided Feature Alignment~\cite{yan2019attention,chen2020deep,liu2021adnet} and Image Translation-Based Alignment~\cite{ks2019deep,lee2020exposure}. After alignment, the process of fusing LDR images has similarities to many image fusion methods, as summarized in~\cite{kaur2021image}. Motion within the multi-exposure is the key factor that influences the method. Messikommer \etal~\cite{messikommer2022multi} utilize the inter-frame motion information provided by the events to assist in the alignment of LDR stacks. While such an approach results in high-confidence light and dark details, multi-frame alignment still produces excessive alignment artifacts for large-scale motion.

\PAR{Deep single-exposure HDR.}~Single-exposure HDR image reconstruction aims to recover missing information in the saturated areas of an LDR image and reconstruct an HDR image using a neural network\cite{wang2021deep}. Representative methods in this domain include HDRCNN\cite{eilertsen2017hdr} and deep reverse tone mapping\cite{endo2017deep}. While the single-exposure HDR approach avoids the alignment problem in bracketing, its recovery of overexposed and underexposed areas is an ill-posed problem, and synthetic artifacts may compromise the credibility of the result.


\section{Event-Based Temporal-Mapping Photography}
\label{sec:methods}

The independent pixels on the event cameras can respond to changes in the brightness $L(x,y,t)$.
\cref{dvs_principle} shows how the event camera works:
\begin{equation}
\label{dvs_principle}
p{(x,y,t_i)}=\left\{
\begin{aligned}
+1 & ,\quad if& L(x,y,t_i)-L(x,y,t_{i-1})>C,\\
-1 & ,\quad if& L(x,y,t_i)-L(x,y,t_{i-1})<-C,
\end{aligned}
\right.
\end{equation}
where $L(x,y,t_i)$ denotes the brightness on the pixel coordinates $(x,y)$ at the current moment $t_i$, $L(x,y,t_{i-1})$ denotes the brightness when the event is last triggered, and $C$ denotes the contrast threshold. When the brightness change exceeds the threshold, events with different polarities are emitted. Brightness is a logarithmic map of light intensity~\cite{gallego2020event}. Intensity images cannot be obtained solely through DVS. Previous work~\cite{rebecq2019high} recovered grayscale frames/videos from moving cameras or in dynamic scenes by training an event-to-video network. However, events cannot be directly integrated to recover accurate grayscale images if both the camera and the scene are static.



\subsection{Principles of Event-Based Temporal-Mapping Photography}
\label{sec:principles}
Returning to the fundamental principle introduced in \cref{dvs_principle}, the core of the event camera is the change in brightness.
Various factors contribute to changes in brightness, with motion in a textured scene being one of them. E2VID~\cite{rebecq2019high} leverages such events to recover its grayscale images. Rather than confining our thinking to motion events, can we focus on the essence of event cameras—detecting changes in brightness?

Suppose that the initial brightness on the DVS sensor plane is 0, namely $L(x, y, t_{i-1})$ is 0 in \cref{dvs_principle}. As we gradually increase the transmittance of the DVS, the brightness on the sensor plane increases. The \ipe of each pixel triggers when $L(x, y, t_{i})$ reaches $C$. Since the corresponding objects in the scene have different intensities, different pixels exhibit distinct rates of brightness growth. Consequently, the brightness $L(x, y, t_{i})$ reaches the threshold $C$ at different times for different pixels. 
Unlike conventional cameras that accumulate photons over a fixed period to measure light intensity, our method incrementally elevates the brightness everywhere on the sensor and records the time reaching a predefined brightness level, \ie, the \ipe timestamp. IPE timestamps are negatively correlated with the absolute light intensity in the scene.
\begin{figure}[t]
    \centering
    \includegraphics[width=0.95\textwidth]{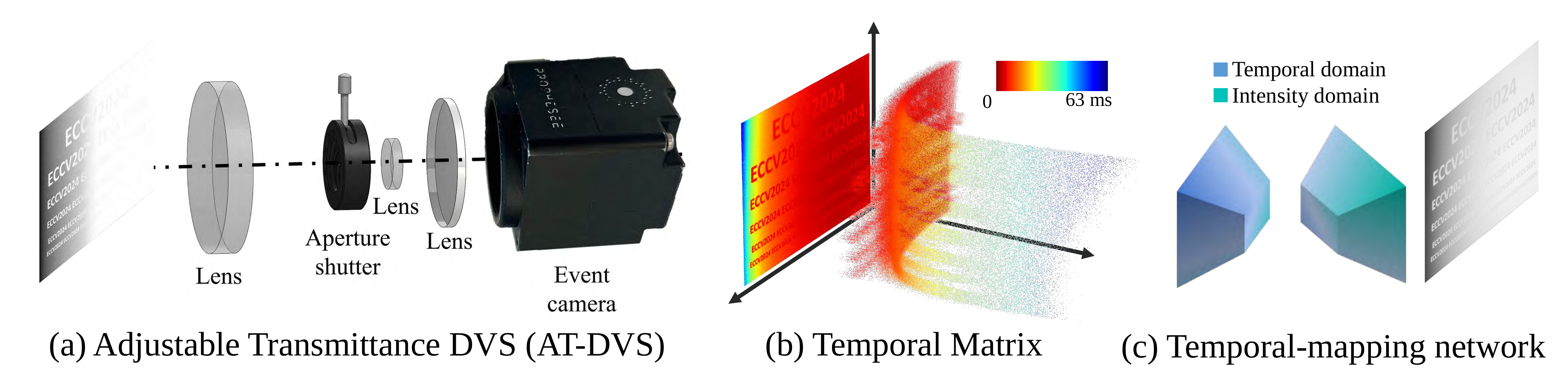}
    \caption{\textbf{Pipeline of \shortname.} The actual scenes are initially recorded by \ourhardware, generating a \timeimage. Following this, the \timeimage undergoes time-intensity mapping and denoising through a Temporal-mapping network, producing a grayscale image.}
    \label{fig:pipeline}
\end{figure}

\subsection{Details of Implementation}
\label{sec:implementation}

\begin{figure}[t]
	\centering
	\begin{subfigure}{0.136\textwidth}
		\centering
		\includegraphics[height=2.4cm]{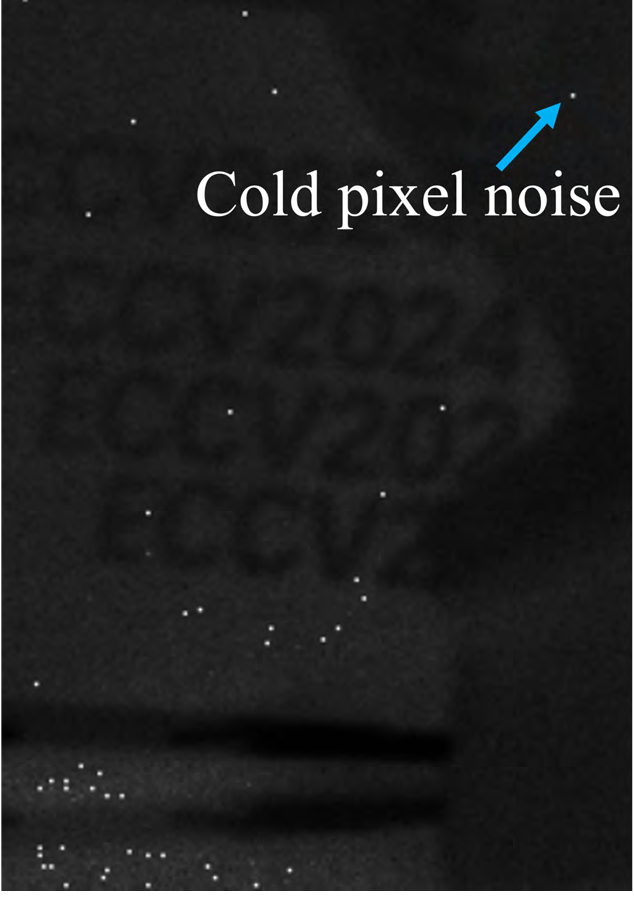}
            \caption{}
		\label{fig:implement_timeimg}
	\end{subfigure}
	\centering
	\begin{subfigure}{0.136\textwidth}
		\centering
		\includegraphics[height=2.4cm]{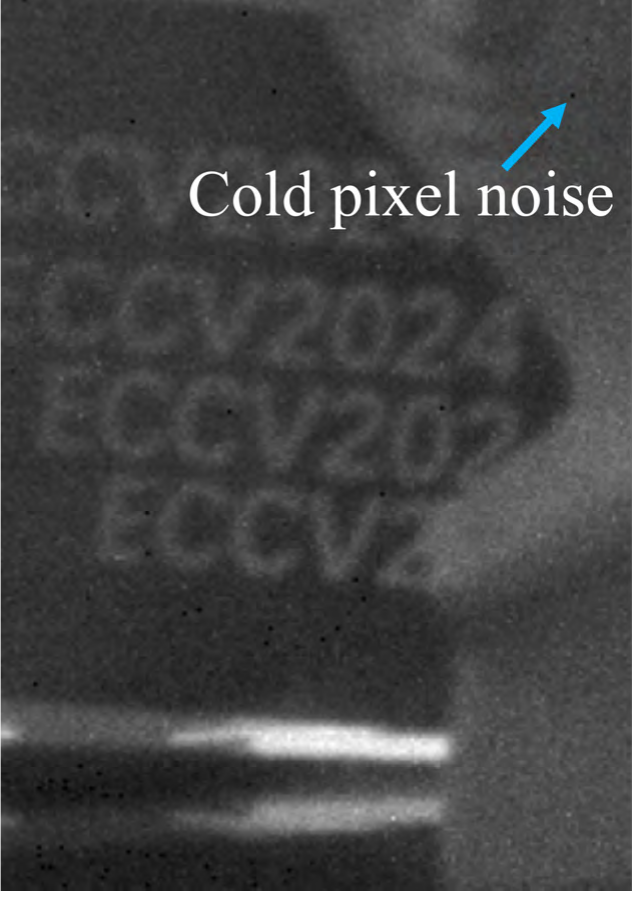}
            \caption{}
		\label{fig:implement_grayimg}
	\end{subfigure}
	\centering
	\begin{subfigure}{0.344\textwidth}
		\centering
		\includegraphics[height=2.4cm]{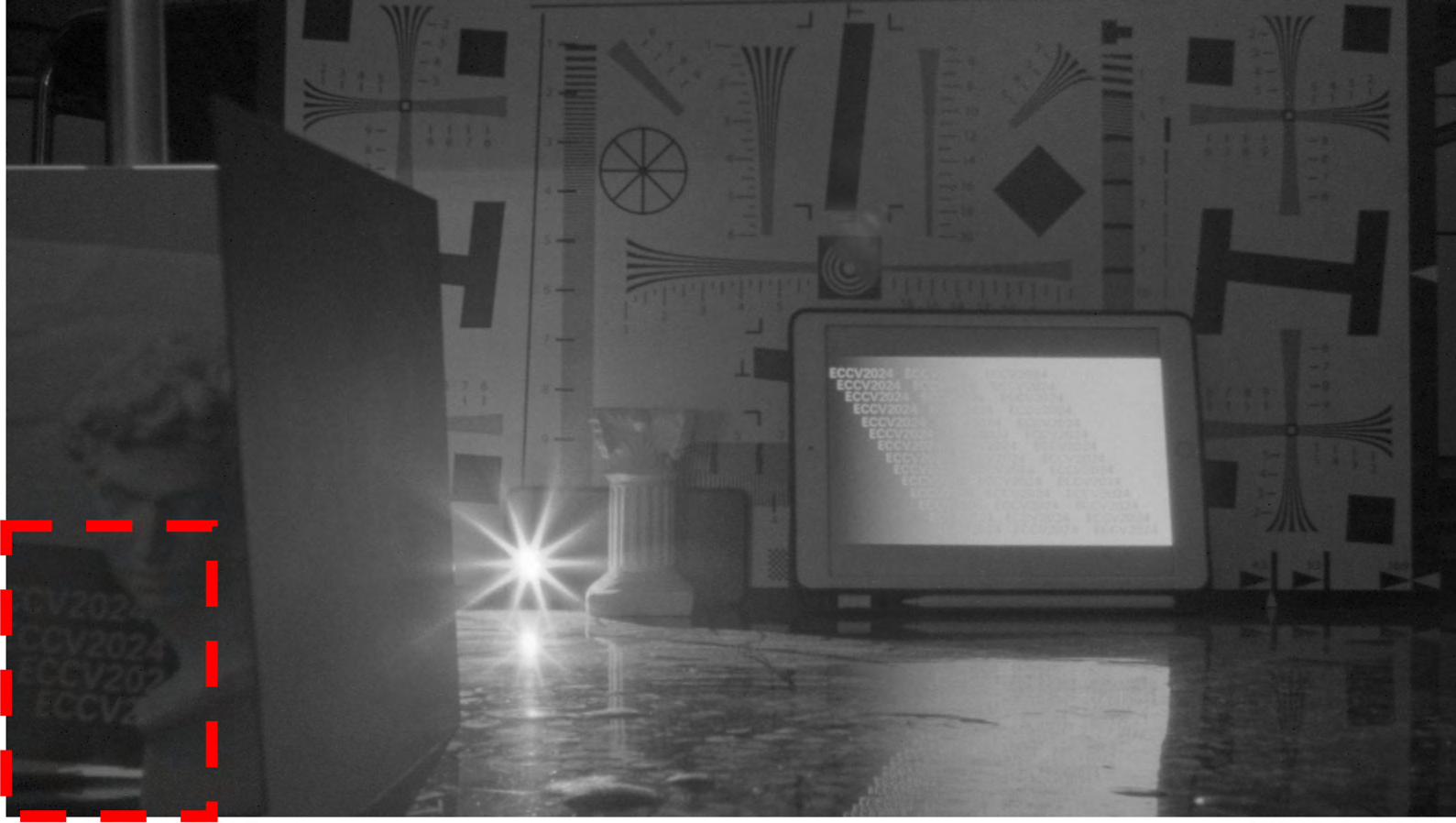}
            \caption{}
		\label{fig:implement_hdr}
	\end{subfigure}
        \begin{subfigure}{0.344\textwidth}
		\centering
		\includegraphics[height=2.4cm]{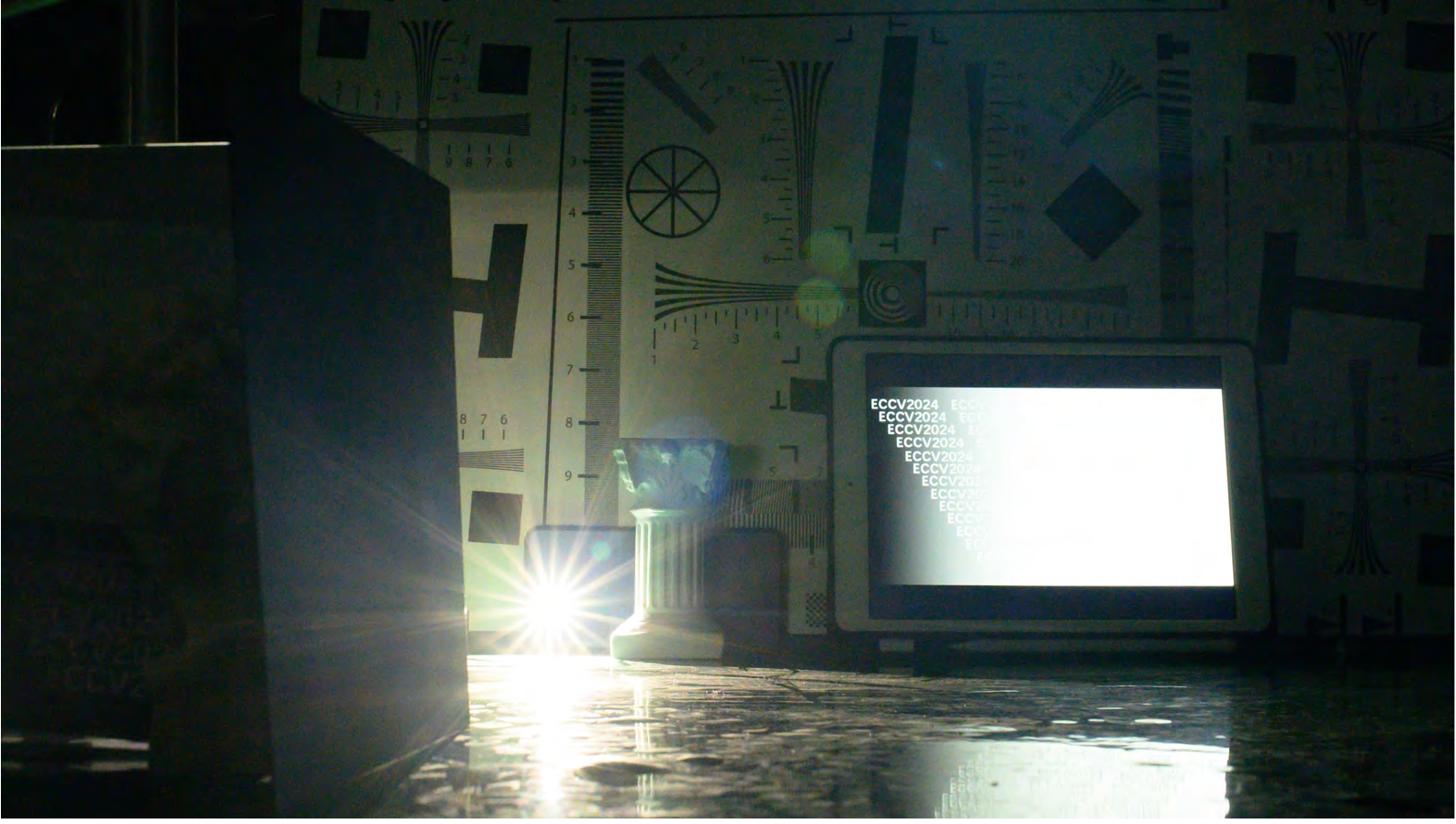}
            \caption{}
		\label{fig:implement_ldr}
	\end{subfigure}
	\caption{\textbf{An implementation example of \shortname.} (a) \timeimage. (b)The intensity image obtained by substituting \timeimage into \cref{t2image}. (c) The adaptive dynamic range image ensures optimal grayscale resolution without underexposure or overexposure. (d) Restricted dynamic range image shows overexposing and underexposing.}
	\label{fig:image_map}
\end{figure}

The transmittance adjustment device (TA) is the key to our method, which needs to be able to adjust the transmittance of the incident light \textbf{starting from 0}. We adopt an aperture shutter as the TA, as shown in \cref{fig:pipeline}a.  Other alternative TAs such as rotary polarizer attenuators and liquid crystal shutters are discussed in the supplementary material.
TA makes the transmittance rate (TR) of the optical system change from 0 to 1. In general, TR changes over time according to the following function:
\begin{equation}
\label{transmittance}
TR(t)=\left\{
\begin{aligned}
f(t) & , & 0 \leq t \leq t_{end}, \\
1 & , & t > t_{end},
\end{aligned}
\right.
\end{equation}
where $f(t)$ is a monotonically increasing function with a value range between 0 and 1, defined by users. $t_{end}$ denotes the point in time when the transmittance reaches 1. By adjusting $t_{end}$, we can control the speed of change of the transmittance, thus affecting the acquisition time of \shortname.

In \shortname, the accuracy of the temporal measurement of the DVS affects the accuracy of the grayscale recovery. Event cameras exhibit significant event delay in low light, where the timestamp of an event lags behind its actual time of occurrence. This delay is due to the decrease in the bandwidth of the photodiode in low light conditions~\cite{posch2010qvga}. In low-light situations, due to the presence of event delays, the $L(x, y, t_{i})$ in \cref{dvs_principle} is no longer the current instantaneous value but should be considered as an integral value from 0 up to the present moment. We interpret \cref{dvs_principle} in the \shortname process  at the circuit level:

\begin{equation}
\label{low-light principle}
V_{diff}=\frac{1}{C_{PD}}\log(1+\int_0^{t^*(x,y)}I_{max}(x,y)\cdot TR(t) \text{d}t) - V_{ref}  \equiv V_{thd},
\end{equation}
where $t^*(x,y)$ is the timestamp when the \ipe is generated for a given pixel $(x,y)$. $C_{PD}$ is the capacitance of the photodiode. $V_{ref}$ is the reference voltage, and since transmittance increases from 0, $V_{ref}$ is ideally 0. $V_{thd}$ is the voltage threshold, quantitatively related to the event trigger threshold $C$ in \cref{dvs_principle}. To conform to the dimension, $I_{max}(x,y)$ can be considered here as the photocurrent when the transmittance is 1. 

As $TR(t)$ increases from 0 to 1, the brightness on the sensor plane increases, and the photocurrent increases accordingly. During this process, the photodiode continues to accumulate photons until $V_{diff}$ exceeds $V_{thd}$. There is a linear mapping relationship between the photocurrent, light intensity, and the grayscale, so $I_{max}(x,y)$ is the target value. 
We perform a simple transformation of \cref{low-light principle} to reflect the mapping from $t^*(x,y)$ to $I_{max}(x,y)$ more clearly: 


\begin{equation}
\label{t2image}
I_{max}(x,y) = \frac{\exp((V_{ref} + V_{thd}) \cdot {C_{PD}})-1}{h(t^*(x,y))}, \\
h(t^*) =\int_0^{t^*(x,y)} TR(t) \text{d}t.
\end{equation}

For grayscale images, we care about the relative values among $I_{max}(x,y)$ rather than their exact magnitudes. So the numerator term, as a coordinate-independent constant, can be eliminated by normalization operations. By substituting each $t^*(x,y)$ in the \timeimage into \cref{t2image} and normalizing $I_{max}(x,y)$ by the maximum and minimum values, we derive the normalized grayscale image $\overline{I_{max}}$, shown in \cref{fig:implement_grayimg}, where $\overline{I_{max}} \in [0,1]^{W\times H}$.

In \cref{t2image}, we define the definite integral over $TR(t)$ as another function $h(t^*(x,y))$ concerning $t^*(x,y)$, which is also a monotonically increasing function. While the true $TR(t)$ may deviate from the predefined function (see \cref{transmittance}), this disparity influences the mapping curve of $\overline{I_{max}}$, akin to the Look Up Table (LUT) mapping curve in traditional images. 

\PAR{Overview.}~\Cref{fig:pipeline} illustrates the pipeline of our work.
First, we integrate the transmittance adjustment device and the event camera, forming \ourhardware. 
Then, we record the events generated when the transmittance increases from 0 and record the timestamp when each pixel triggers the \ipe, forming a \timeimage. Subsequently, through a Temporal-mapping network, we perform time-intensity mapping and denoising to the \timeimage. And finally, we get the result of \shortname. 

\PAR{Adaptive Dynamic Range.}~For conventional cameras, the dynamic range of is a fixed value (typically 70 dB) when the exposure parameters are fixed. 
In high dynamic range scenes (103 dB in the scene of \cref{fig:implement_hdr}), single-exposure images from conventional cameras are often overexposed and underexposed, as presented in \cref{fig:implement_ldr}.
However, \shortname enables adaptive dynamic range photography, effectively preserving both dark and light details within the image. This capability stems from the fact that the \timeimage operates within the range spanning from the brightest pixel $t^*$ to the darkest pixel $t^*$.


\PAR{High grayscale resolution.}~Another benefit of \shortname is its higher gray-scale resolution.  
For conventional cameras, a 12-bit raw format file is capable of storing up to 4096 gray levels.
In contrast, our \shortname utilizes the high temporal resolution of the event camera to achieve extremely high grayscale resolution. Taking the scene in \cref{fig:implement_hdr} as an example, the difference in $t^*$ between the brightest and darkest is 292111 $\mu$s, and the temporal resolution of the event camera is 1 $\mu$s. Theoretically, the effective gray resolution level of \cref{fig:implement_hdr} is 292111 levels, which is \textbf{71.3 times} that of a 12-bit raw format file.

\subsection{Degradation Model for EvTemMap}
\label{sec:degradation_model}


\Cref{t2image} delineates the ideal formulation for the conversion of events to intensity image. However, for event cameras in the real world, there are many practical factors that make it impossible to derive high-quality intensity images directly from the events: (1) uncertain threshold $C$, (2) event timestamp error, (3) noise brought by cold and hot pixels, and (4) other degradations.
Previous work~\cite{wang2021real,zhang2021designing} designed a practical degradation model to restore low-quality images. Inspired by this, we analyze each factor for our temporal-mapping imaging in both \textbf{temporal domain} and \textbf{spatial domain} and design a degradation space as large and realistic as possible for our \shortname.

\begin{figure}[t]
    \centering
    \includegraphics[width=0.95\textwidth]{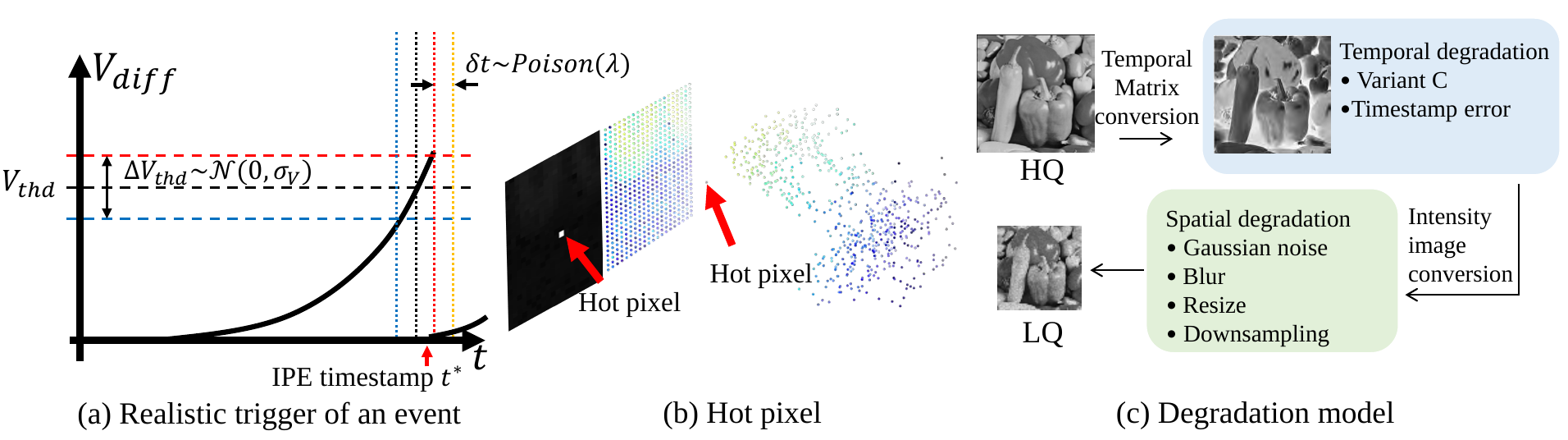}
    \caption{\textbf{Degradation paradigm.} (a) The real \ipe timestamp is affected by the threshold random perturbation $\Delta V_{thd}$ and the time measurement error $\delta t$. (b) The hot pixel is reflected as an unusually early value on the \timeimage and as the brightest white dot in the grayscale image. (c) The process of generating training data for the degradation model.}
    \label{fig:degradation}
\end{figure}

\PAR{Spatially variant contrast threshold.}~The threshold values $C$ for a given event camera vary both over time and across the image instead of remaining constant~\cite{stoffregen2020reducing}, which results from the variant $V_{thd}$ in \cref{t2image}, as shown in \cref{fig:degradation}a.
In addition, $V_{ref}$ is not ideally 0 due to dark current or TA light leakage.
We consider the term $((V_{ref} + V_{thd}) \cdot {C_{PD}})$ as the threshold $C$ for the event camera and summarize the above sources of noise, embodied as random perturbations of $C$. 
To cope with different threshold conditions, we choose $C$ randomly from 0.1 to 0.4 for each sample.
Then we add a $\Delta C$ to the set threshold $C$ for different pixels, where the $\Delta C$ obeys the zero-mean Gaussian noise model $\mathcal{N}(0,\mathbf{\sigma})$. $\mathbf{\sigma}$ is uniformly sampled from 0.02 to 0.08.


\PAR{Event timestamp error} comes from the time uncertainty of DVS output events. This time uncertainty is manifested as a time delay that approximates a Poisson distribution~\cite{hu2021v2e}. This temporal measurement noise is also spatially random on \shortname images. 
In our degradation model, we add $\delta t$ to $t^*(x,y)$ in \cref{t2image}, and $t$ is sampled from $\text{Poisson}(\lambda)$, where $\lambda$ is uniformly sampled from 10 to 1000 $\mu\text{s}$. Extreme conditions are discussed in supplementary materials.




\PAR{Cold and hot pixels.}~Because of the limitation of the circuit, there are cold pixels, which are less prone to generating events, and hot pixels, which constantly generate events in DVS~\cite{hu2021v2e}.
In the \timeimage, the cold pixel is the location where $t^*$ is null, which is artificially set to the maximum value of $t^*$, as shown in \cref{fig:implement_timeimg,fig:implement_grayimg}. In contrast, the hot pixel exhibits an unusually early $t^*$, resulting in the brightest white dots in the grayscale image, as shown in \cref{fig:degradation}b.
Experimentally, it is found that hot pixels consistently occupy specific locations. 
We locate the position of hot pixels with pixel calibration~\cite{Wang2020EventCC}.
We then remove the hot pixels by defining the $t^*$ of them as the minimum of the other positions in \timeimage, \ie temporal pre-processing.



\PAR{Other degradation.}~Apart from the above degradation, there is still some unpredictable degradation such as out-of-focus blur, noise from dark current, \etc. Thus we choose Gaussian noise model $\mathcal{N}(\mathbf{\mu},\mathbf{\sigma})$, where $\sigma$ is uniformly sampled from 1/255 to 25/255. For the out-of-focus blur, we add the same blur kernel as BSRGAN~\cite{zhang2021designing}. The current event cameras suffer from relatively low resolution due to limitations in transmission bandwidth~\cite{gehrig2022high}. To manifest the need for high resolution, we super-resolve the final result of \shortname with a scale factor of 4. Thus we also incorporate downsampling degradation into our degradation space following~\cite{zhang2021designing}.

Up to this point, we have thoroughly analyzed various types of degradation within the framework of \shortname. To enlarge the degradation space, we use the random shuffle strategy for all the degradation processes~\cite {zhang2021designing}.
\cref{fig:degradation}c shows our degradation model. In the training stage, given a high-quality (HQ) image, we first convert the image from the spatial domain to the temporal domain with \cref{t2image} and apply degradation for variant contrast threshold, event timestamp error, and cold pixels. After the temporal degradation, we convert the degraded Temporal Matrix to intensity image and apply the rest spatial degradation. Finally, the resulting low-quality image (LQ) is used as the input of the Temporal-mapping network for training. In contrast to E2VID~\cite{rebecq2019high}, our approach does not require synthetic events for training, thereby obviating the need to consider the domain gap issue arising from it.

\section{Experiments}
\label{sec:experiments}

\subsection{TemMat Dataset}
For comparison, we collect \ourdataset. The dataset features (1) diverse scenarios including both outdoor and indoor scenes, (2) a significant luminance range from 0.02 lux to 20000 lux, and (3) high dynamic range scenes, reaching up to 114 dB. We collect the Temporal Matrix with AT-DVS. To compare with other E2VID~\cite{rebecq2019high,stoffregen2020reducing} methods, for each scene, we also collect motion events by slowly and gently shaking the DVS, without changing the transmittance of the optical system. For capturing reference images, a Nikon D780 SLR camera equipped with an AF-S NIKKOR 50mm f/1.4G lens, configured in bracket-exposure HDR mode, is employed. To achieve a depth of field comparable to that of the event camera, an aperture of F16 is utilized in aperture priority mode. Because the Temporal Matrix, motion events for E2VID, and HDR images by SLR camera are obtained separately, the results of \shortname, E2VID, and reference images are not pixel-aligned.

\subsection{Settings}
\label{sec:settings}
\PAR{AT-DVS details.}~We adopt prophesee EVK4 as DVS sensor in our experiments, with a spatial resolution of 1280$\times$720 and a temporal resolution of 1 $\mu\text{s}$. Different from DAVIS, which outputs spatially aligned intensity images and events, EVK4 only outputs events. For the transmittance function $TR(t)$, we place an aperture shutter at the aperture diaphragm to control the transmittance of the optical system by adjusting the aperture. Please refer to the supplement for more details on our AT-DVS.

\PAR{Training details.}~The novelty of the paper lies in the brand new event-to-image conversion method, \ie temporal mapping and corresponding degradation model. To demonstrate the advantage of the proposed method, we choose widely-used Swinir~\cite{liang2021swinir} as our image restoration model for the temporal-mapping network. The model is trained on a composition comprising DIV2K~~\cite{agustsson2017ntire}, Flick2K~\cite{lim2017enhanced,timofte2017ntire}, WED~\cite{ma2016waterloo}. With the HQ images in the dataset, we produce the LQ images with the proposed temporal and spatial degradation model on the fly. We trained model with 72 $\times$ 72 LQ patch size. Following BSRGAN~\cite{zhang2021designing}, we train the model with a weighted combination of L1 loss, perceptual loss~\cite{johnson2016perceptual}, and spectral norm-based least square PatchGAN loss~\cite{isola2017image} with weights 1, 1, and 0.1, respectively. We train the model using Adam~\cite{kingma2014adam} with an initial learning rate of 0.0002 and cosine learning rate decay with a minimum learning rate of $10^{-7}$. We train the model with a batch size of 48 for 1000k iterations.

\PAR{Compared methods.}~We are the first method that is able to convert a static scene from a stationary event camera to the best of our knowledge, the most similar work is E2VID, which translates motion events to video. Thus, we compare our method with E2VID~\cite{rebecq2019high} and E2VID+~\cite{stoffregen2020reducing}.




\subsection{Comparisons with State-of-the-Art Methods}

\begin{figure}[t]
  \centering
  \includegraphics[width=0.99\linewidth]{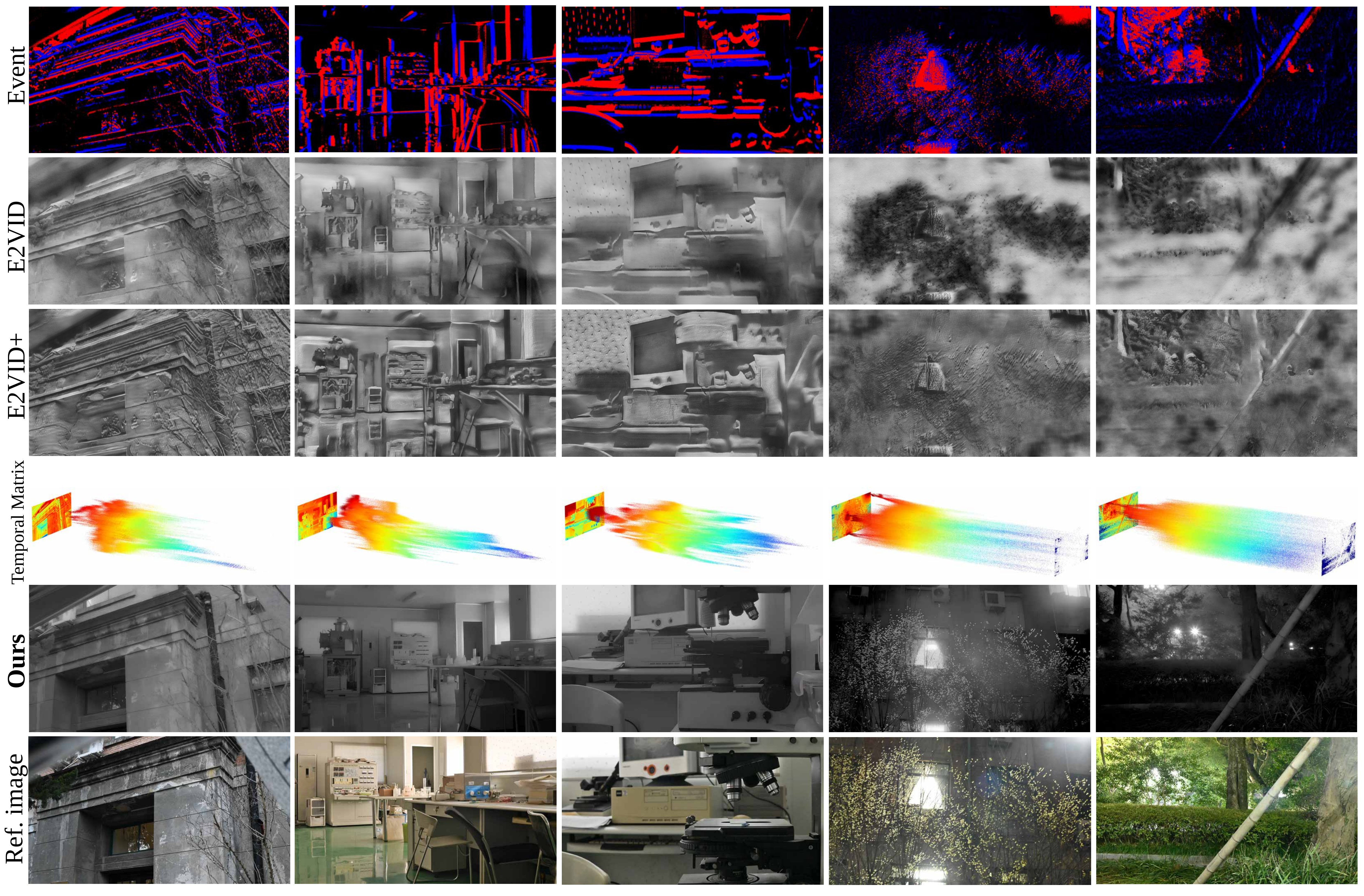}
  \caption{\textbf{Qualitative results on \ourdataset.} Rows from top to bottom: Visualized event frame, results of E2VID~\cite{rebecq2019events} and E2VID+~\cite{stoffregen2020reducing}, visualized \timeimage, results of \shortname, reference image. Our method achieves the most faithful and finest result. Best viewed on a screen and zoomed in.}
  \label{fig:qualitative}
\end{figure}

\PAR{Qualitative results.}
Qualitative results on the \ourdataset are presented in \cref{fig:qualitative}, comparing our approach with two event-to-video methods: E2VID~\cite{rebecq2019events} and E2VID+~\cite{stoffregen2020reducing}. 
Our approach achieves grayscale image recovery closest to the reference, showcasing the realism in our grayscale restoration in both indoor and outdoor scenes.
Specifically, in the leftmost column, our method effectively captures the grayscale transition from outdoors to indoors, and even reflections from glass are discernible.
In addition, in the night scenes presented in the two right columns, the trees in the shadows and the lights in the bright areas are correctly grayscale mapped.
However, while the E2VID and E2VID+ methods roughly depict the outlines of the scene in the left three columns, their grayscales are significantly distorted. In the last two columns, they recover a grayscale that is even opposed to the real situation.

For high-frequency textures, our \shortname also performs far better than other event-based methods, such as the tree branches and blooming flowers in the fourth column. E2VID and E2VID+, on the other hand, only recover the general outline of the scene but blur the details, such as tree branches. 
Intriguingly, our results showcase the capability to discern the approximate texture of illuminated indoor environments from outdoor settings at night, a challenge even for HDR images from SLR cameras.

\begin{table}[h]
    \caption{Quantitative comparison on \ourdataset.}
    \label{table:Quantitative}
    \centering
    \resizebox{0.9\textwidth}{!}{
    \setlength{\tabcolsep}{10pt}     
    \renewcommand\arraystretch{1.0} 
        \begin{tabular}{ l  ||  c  |  c  |  c |  c } 
		\bottomrule[0.15em]
        \rowcolor{tableHeadGray}
		Methods & \textbf{\shortname} & \textbf{E2VID~\cite{rebecq2019events}} & \textbf{E2VID+~\cite{stoffregen2020reducing}} &        \textbf{SPADE~\cite{cadena2021spade}} \\ \hline \hline
            PSNR & \textbf{15.75} & 10.48 & 11.21 & 10.20 \\
            SSIM & \textbf{0.69} & 0.58 & 0.55 & 0.52 \\  \hline
	\end{tabular}}

	\label{tab:method_comparison}
\end{table}
\PAR{Quantitative results.} \Cref{table:Quantitative} presents the quantitative comparison results with the E2VID-based methods~\cite{rebecq2019events,stoffregen2020reducing,cadena2021spade}. We take the bracketed HDR grayscale images captured by the SLR camera as the ground truth, and use LightGlue~\cite{lindenberger2023lightglue} to align the results of various methods with the ground truth, and then perform PSNR and SSIM evaluations. In terms of quantitative results, \shortname performs the best, while the three E2VID-based methods do not differ much. More detailed quantitative and qualitative results can be explored in the supplementary material.

\subsection{High Dynamic Range Photography}
\label{sec:hdr_experiments}



\begin{figure}[t]
	\centering
	\begin{subfigure}{0.24\linewidth}
		\centering
		\includegraphics[width=\linewidth]{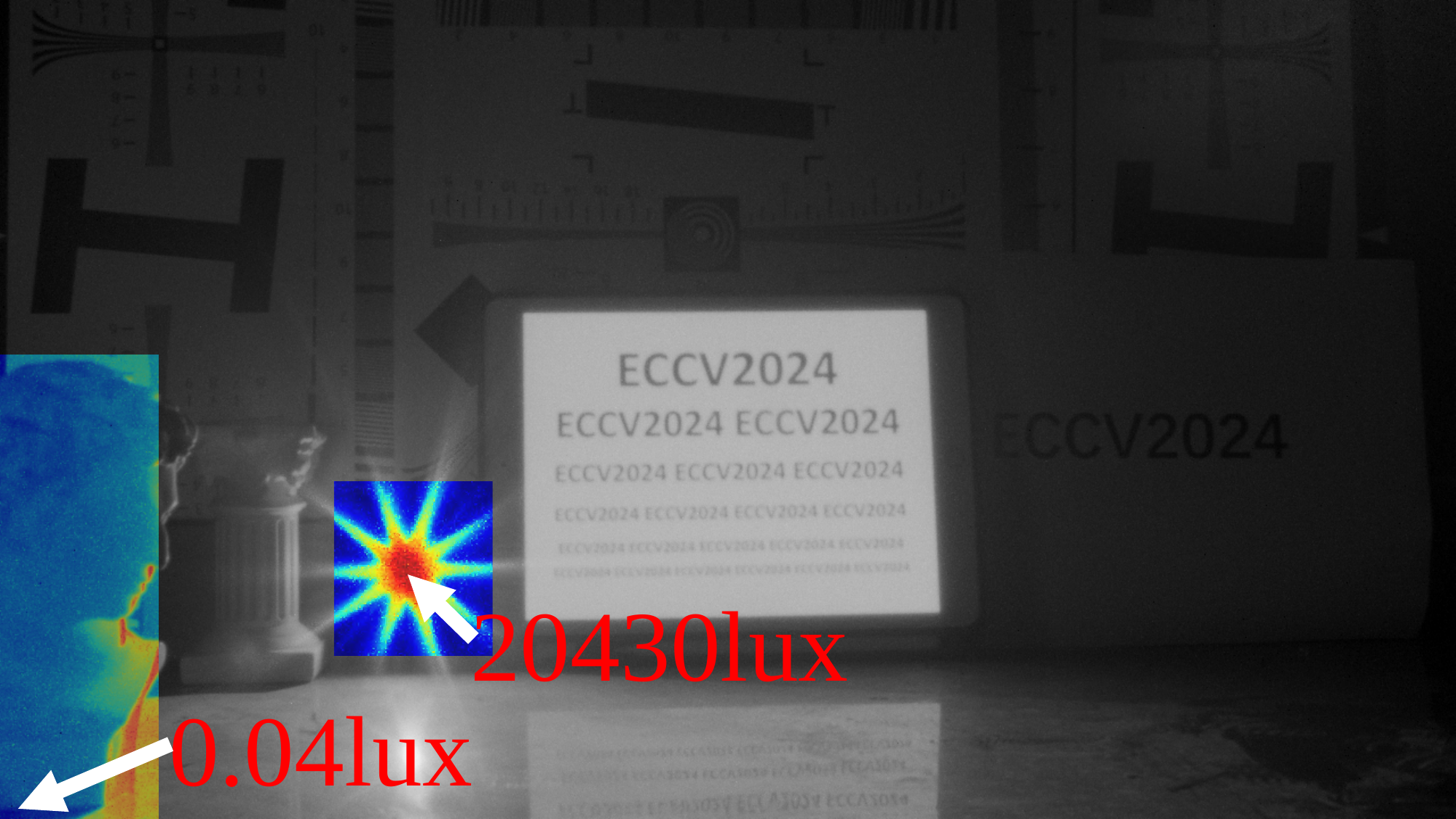}
            \caption{\shortname}
		\label{fig:hdr_evtemmap}
	\end{subfigure}
        \centering    
	\begin{subfigure}{0.24\linewidth}
		\centering
		\includegraphics[width=\linewidth]{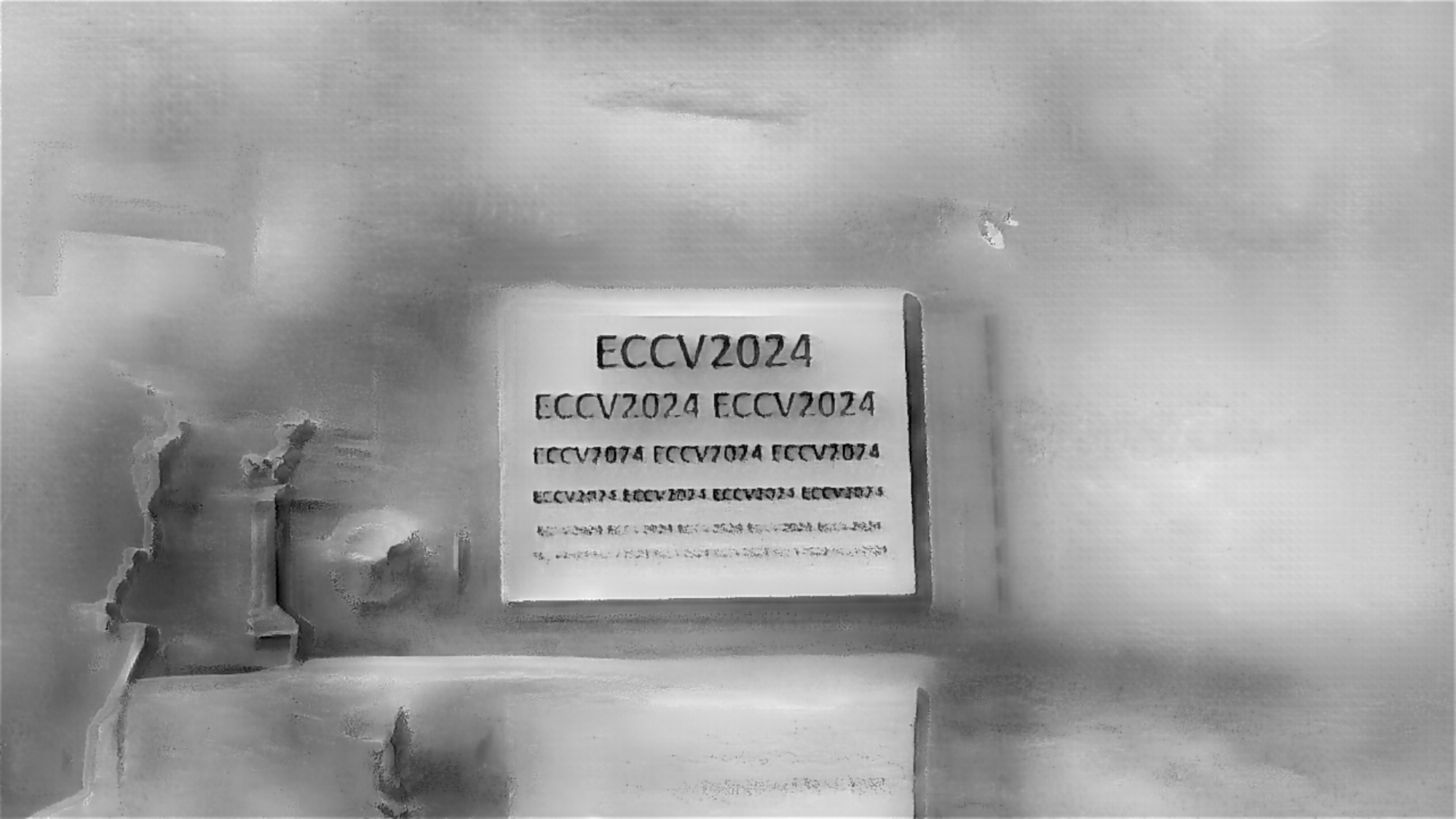}
            \caption{E2VID}
		\label{fig:hdr_e2vid}
	\end{subfigure}
        \begin{subfigure}{0.24\linewidth}
		\centering
		\includegraphics[width=\linewidth]{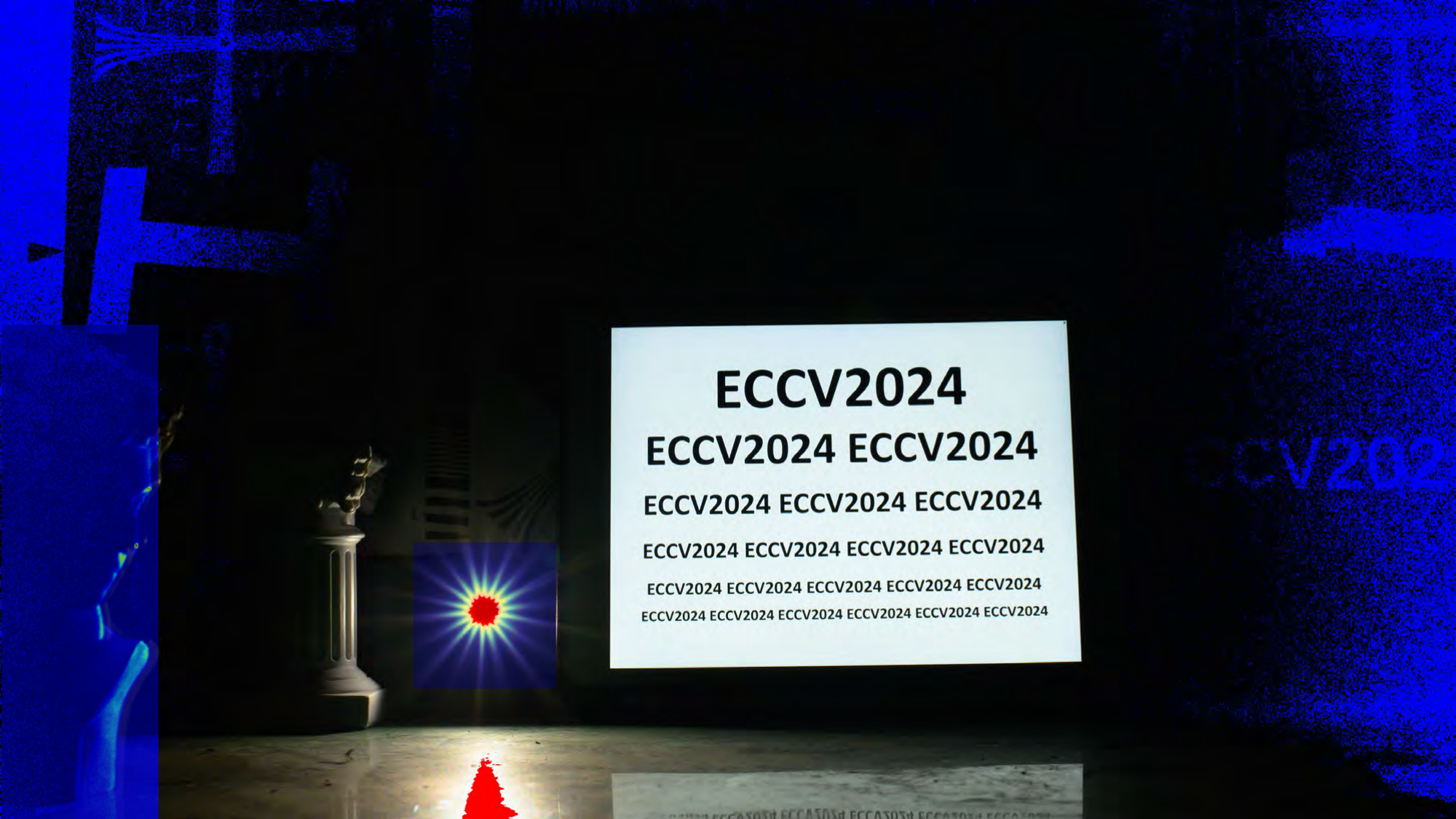}
            \caption{Conventional LDR}
		\label{fig:hdr_ldr}
	\end{subfigure}
	\centering
	\begin{subfigure}{0.24\linewidth}
		\centering
		\includegraphics[width=\linewidth]{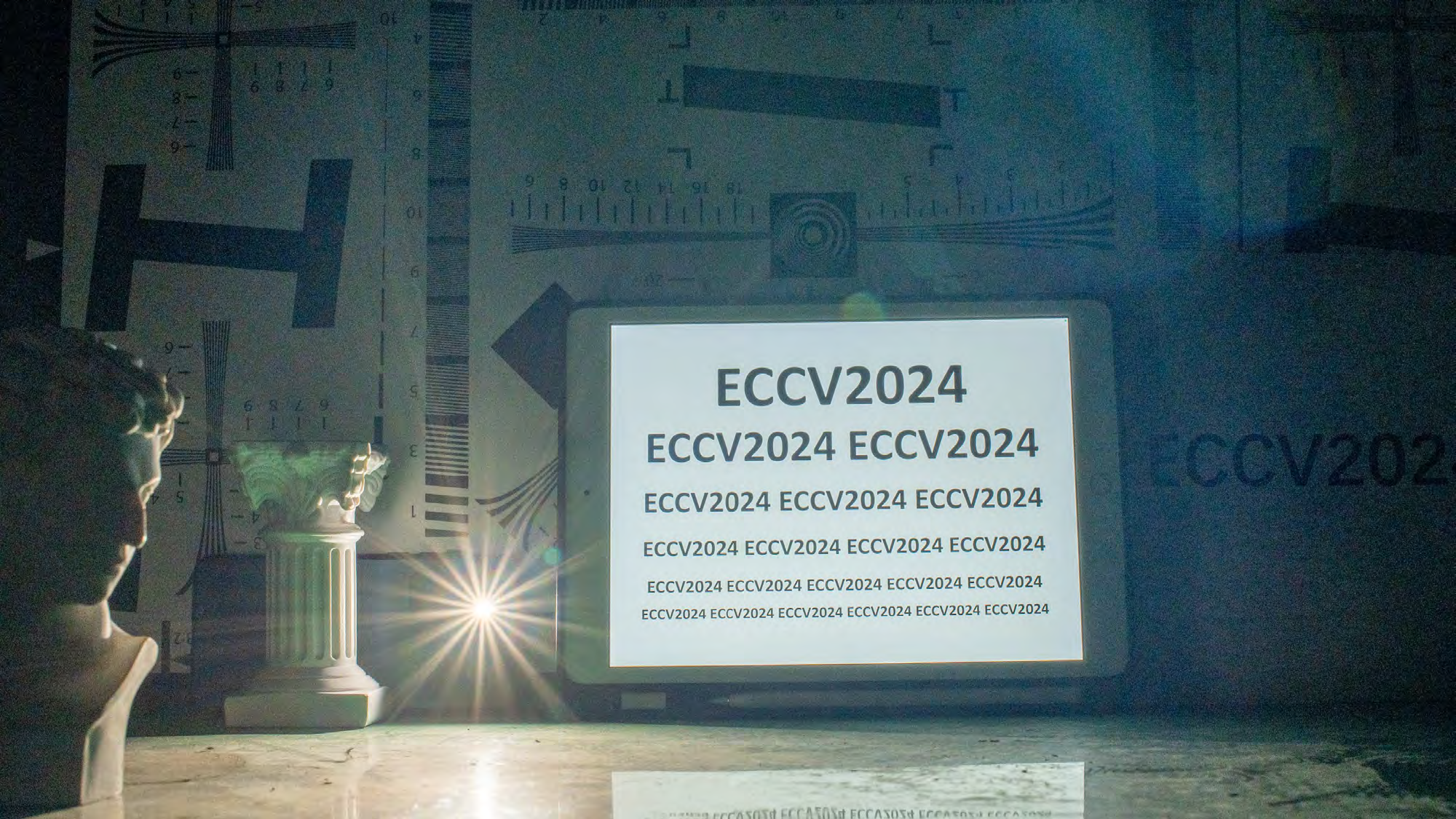}
		\caption{Bracketed HDR}
		\label{fig:hdr_hdr}
	\end{subfigure}
        \caption{\textbf{Comparison in high dynamic range scenes.} The brightest and darkest areas of (a) and (c) are rendered in pseudo-color. Overexposed areas and underexposed areas in (c) are marked in red and blue, respectively. \shortname presents exceptional HDR photography with minimal noise. Conversely, E2VID exhibits a loss of many texture features, while Conventional LDR suffers from inadequate information capture in both underexposed and overexposed regions. }
 %
    \label{fig:HDR}
\end{figure}

To test the dynamic range of our \shortname, we construct an HDR scene with a range of 114dB in a dark room, featuring 20430 lux at the brightest point and 0.04 lux at the darkest point measured by an illuminance meter. \Cref{fig:hdr_evtemmap} shows that \shortname records details in both shadow and highlight areas without overexposure or underexposure.
Around the brightest point, we can observe the gradation of brightness within the elliptical light source as well as the diffracted asterisms around it. Near the darkest point, the result of \shortname effectively captures intricate details in the hair of David's sculpture and natural variations in light and shadow on the shoulder. In the rest of the image, the ``ECCV2024'' texts on the screen and in the shadows, as well as the background resolution board, are clearly visible.
Comparing \cref{fig:hdr_evtemmap} and \cref{fig:hdr_e2vid}, it can be found that \shortname can truly depict various details in the scene, while E2VID can only restore textures with high contrast.
Compared with the single-exposure result in \cref{fig:hdr_ldr}, our method captures more details in both shadow and highlight grayscale variations than the conventional LDR image. The outcome from the conventional camera using bracketed HDR is depicted in \cref{fig:hdr_hdr} for reference.




\subsection{Low-light Photography}
\label{sec:low_light_experiments}


\begin{figure}[t]
	\centering
	\begin{subfigure}{0.24\linewidth}
		\centering
		\includegraphics[width=\linewidth]{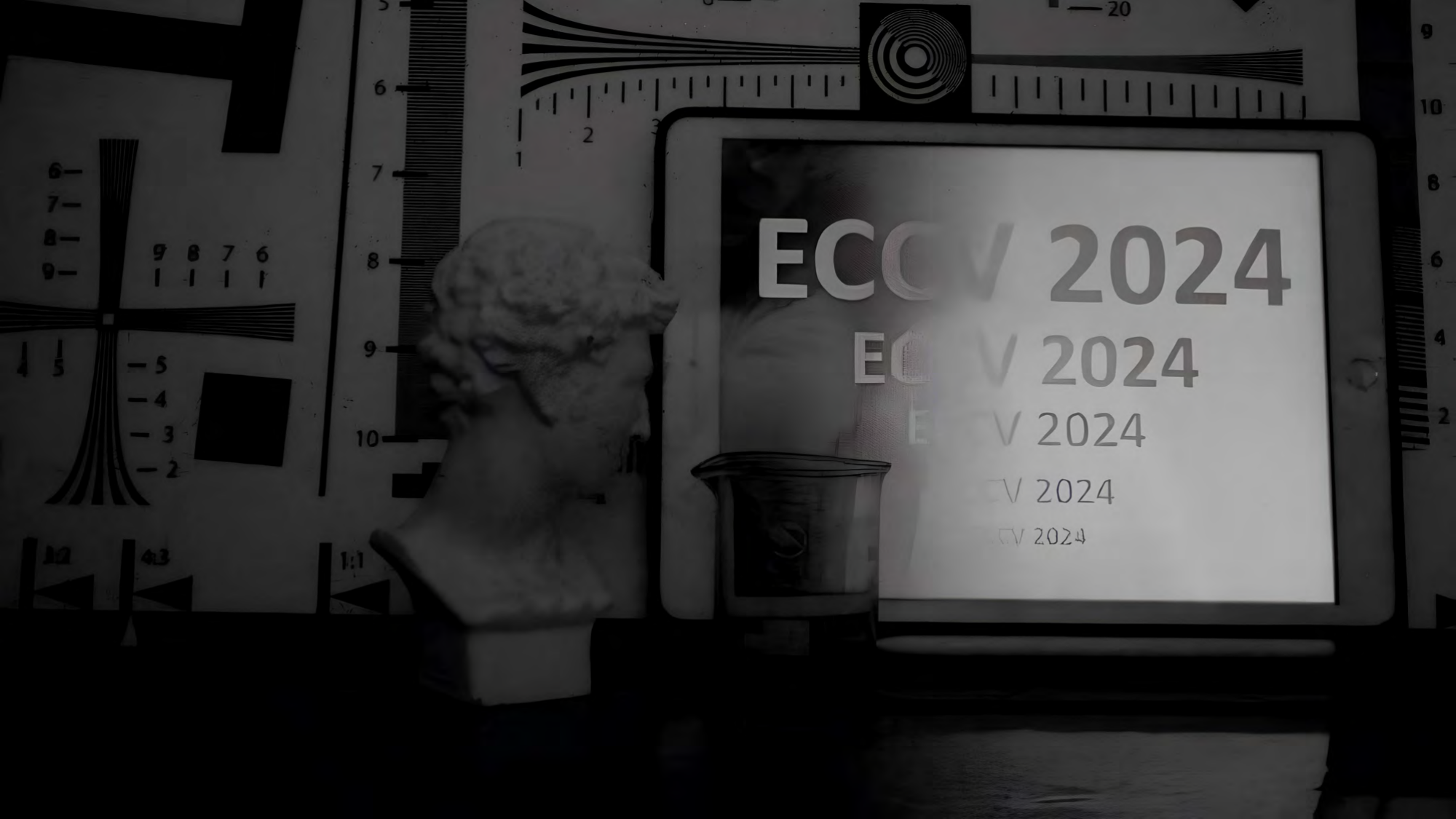}
            \caption{\shortname}
		\label{fig:lowlight_denoised}
	\end{subfigure}
	\centering
	\begin{subfigure}{0.24\linewidth}
		\centering
		\includegraphics[width=\linewidth]{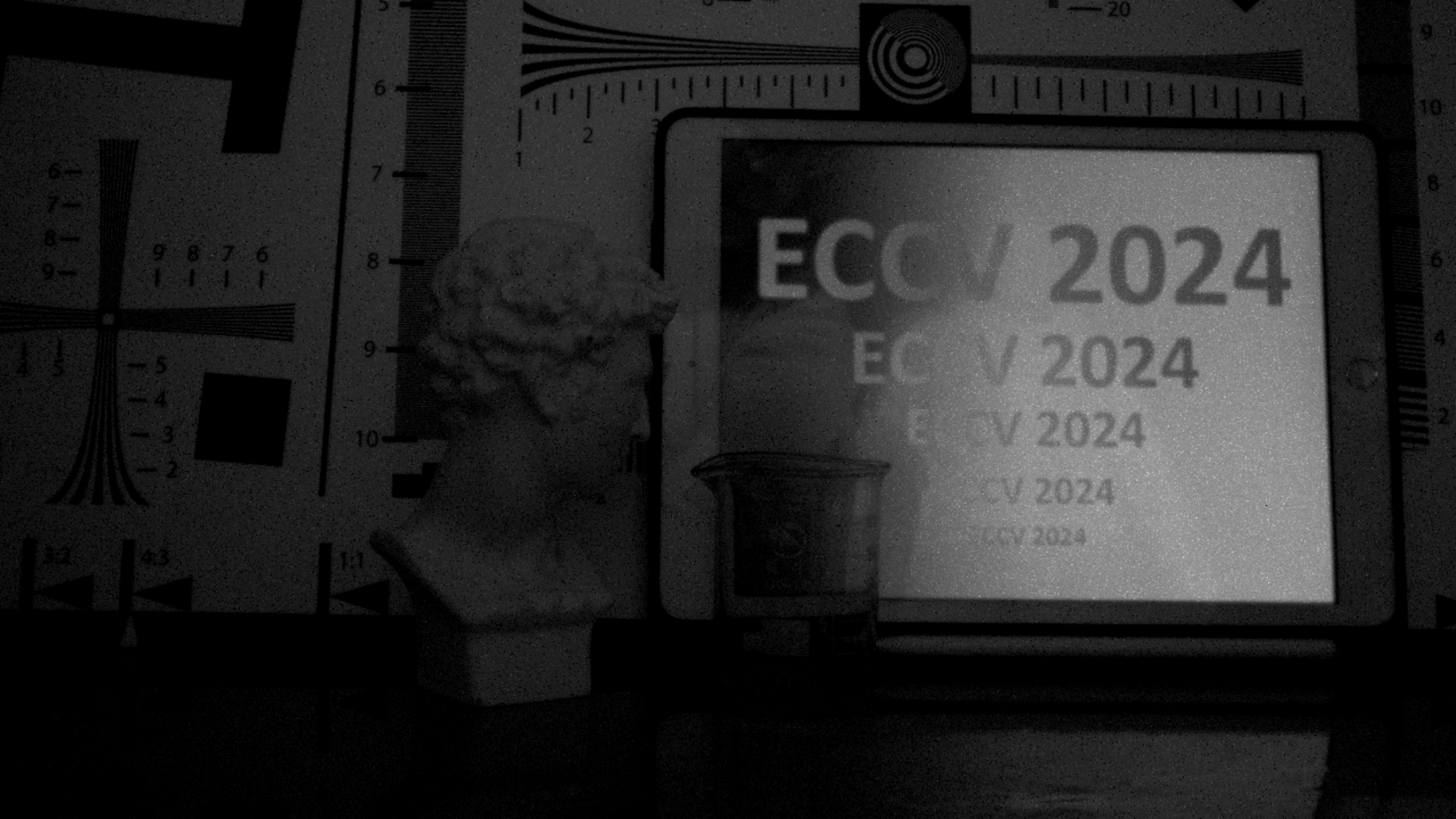}
            \caption{\tiny{Img. from Temp. Mat.}}
		\label{fig:lowlight_raw}
	\end{subfigure}
        \centering    
	\begin{subfigure}{0.24\linewidth}
		\centering
		\includegraphics[width=\linewidth]{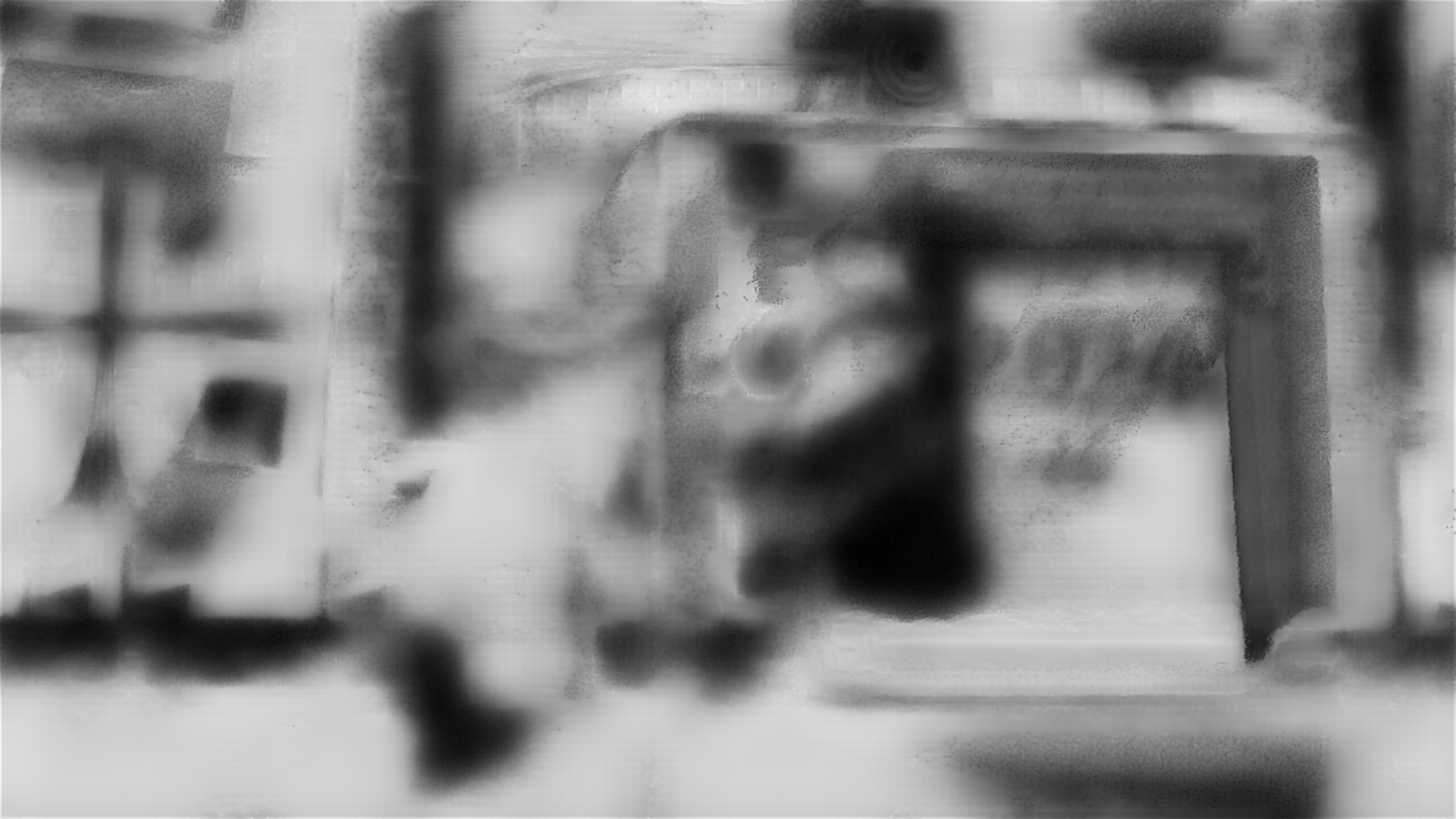}
            \caption{E2VID}
		\label{fig:lowlight_e2vid}
	\end{subfigure}
        \begin{subfigure}{0.24\linewidth}
		\centering
		\includegraphics[width=\linewidth]{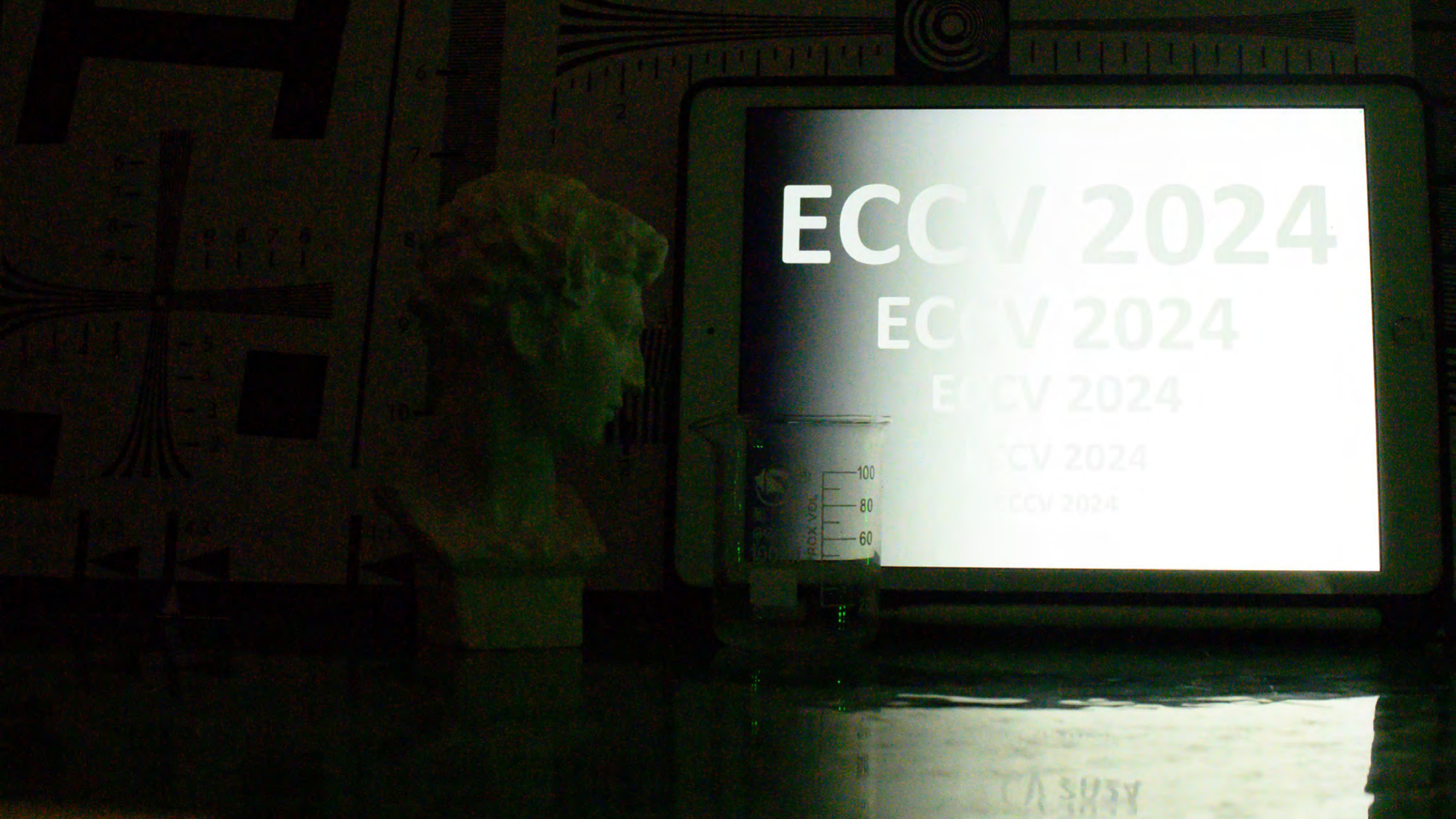}
            \caption{Conventional LDR}
		\label{fig:lowlight_ldr}
	\end{subfigure}
	\caption{\textbf{Comparison in <0.01lux environment.} (a): \shortname shows clear textures and natural grayscale variations. (b): The raw intensity image from \timeimage contains a lot of random noise. (c): E2VID loses a lot of texture. (d): Conventional LDR contains a lot of noise.}
    \label{fig:low_light}
\end{figure}

We challenge the limits of \shortname in extremely dark environments (less than 0.01lux). In \cref{fig:lowlight_denoised}, the texts on the screen and the resolution board line pairs in the dark background are clearly visible. 
Comparing \cref{fig:lowlight_denoised} to \cref{fig:lowlight_raw} highlights the denoising capability of our image restoration model in low-light conditions.
As shown in \cref{fig:lowlight_e2vid}, E2VID presents a worse result in extremely dark environments. It loses much of the texture detail in the scene, making it challenging to locate the target. Compared to \cref{fig:lowlight_ldr}, we highlight the short ``exposure time'' in extremely dark conditions: the conventional LDR image requires an exposure time of up to 1.3 s to capture images in the low-light environment, whereas our AT-DVS only needs 78 ms for the capture of Temporal Matrix, which \textbf{reduces 94\%} of the conventional exposure time. Note that low light condition also brings more noise to conventional LDR results.




\subsection{Motion Photography}
\label{sec:motion}

\begin{figure}[t]
  \centering
  \includegraphics[width=0.96\linewidth]{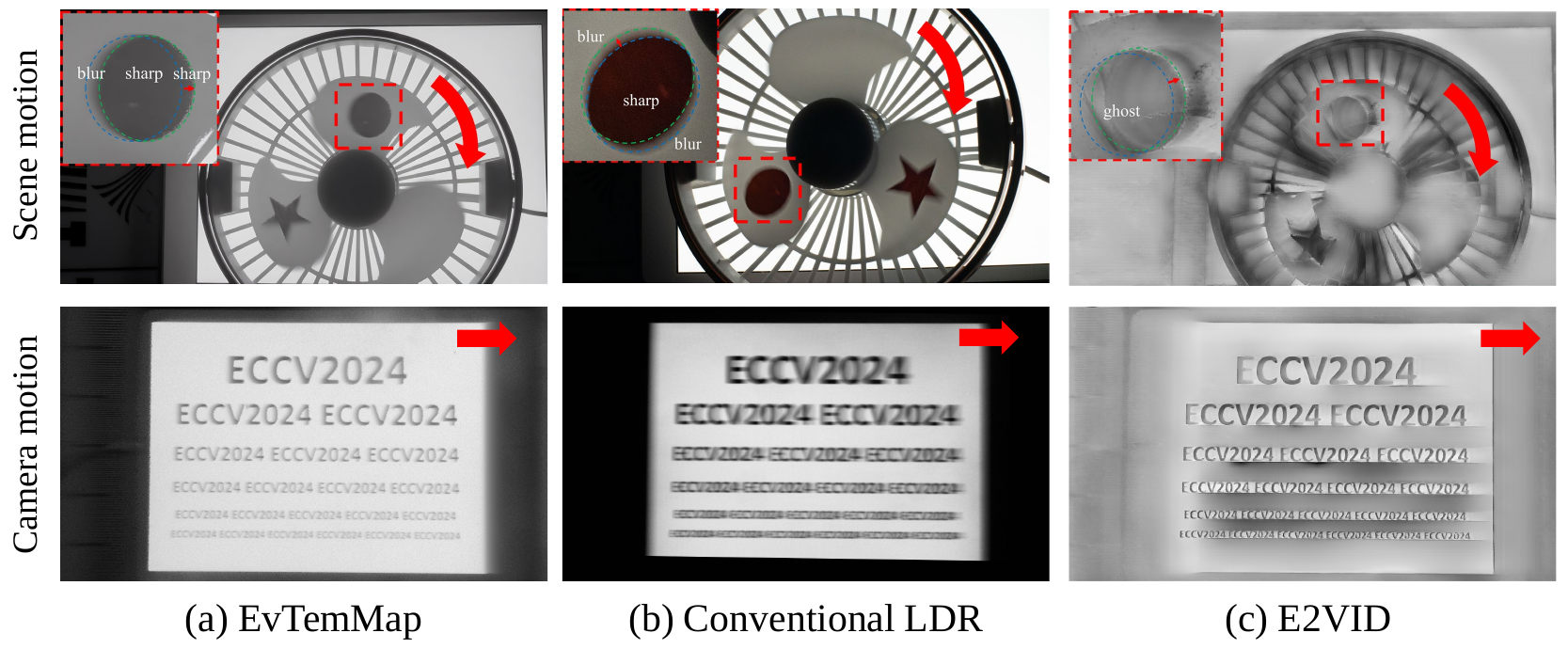}
  \caption{\textbf{Comparison in motion scenes.}(a) Motion blur in \shortname occurs only on \textbf{one} side. (b) Motion blur in conventional image occurs on both sides. (c) Motion blur in E2VID exhibits ``ghosting'' effects.}
  \label{fig:motion}
\end{figure}


Because the time for capturing Temporal Matrix is not ignorable, EvTemMap also experiences motion blur in motion photography, but it presents different characteristics than motion blur in conventional and E2VID images. 

\Cref{fig:motion} shows results from \shortname, a traditional camera and E2VID. The top row shows an example of scene motion, where the fan rotates at a fixed speed. Comparing \shortname images with conventional LDR images, the motion blur in the \shortname image occurs on a single side of the circle during the motion from blue to green, whereas the motion blur in conventional image occurs on both sides of the circle. 
In another example of camera motion, \ourhardware and a conventional SLR camera are combined into a coaxial system with a beam splitter to capture a fixed screen while the system moves. With both cameras set to an exposure time of 40 ms, the bottom row demonstrates that \shortname produces less motion blur compared to the conventional LDR image.
In \cref{fig:motion}c, while E2VID also reconstructs grayscale for motion scenes, it suffers from "ghosting" effects (\cf \cref{sec:e2vid}) and fails to handle static areas of the scene effectively (\eg, the static background in the top row).

\shortname's unique motion blur characteristics arise from its temporal mapping principle. Unlike conventional cameras that use integral imaging to capture the entire image during a fixed integration period, \shortname varies its grayscale acquisition speed based on the brightness of different areas in the image. Brighter areas are captured faster, resulting in less motion blur. Theoretically, \shortname's maximum motion blur occurs in the darkest areas. Although these areas experience the same amount of motion as in conventional images for the same exposure time, the motion blur is only half as prominent because it occurs on only one side, which demonstrates the advantage of \shortname over conventional SLR in motion photography.

\subsection{Ablation Study}
\label{sec:ablation}



\begin{figure}[t]
    \centering
    \includegraphics[width=0.95\linewidth]{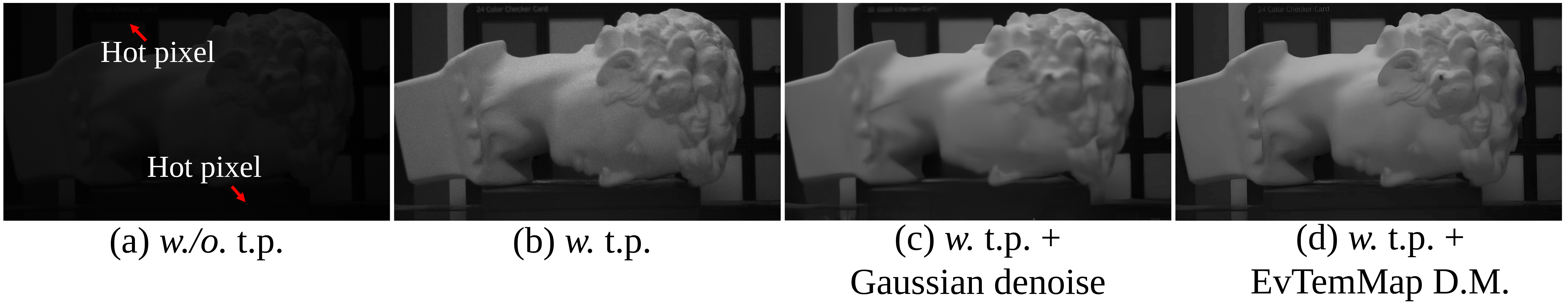}
    \caption{\textbf{Visual results of ablation experiment.} t.p.: temporal pre-processing, D.M.: Degradation Model. (a): The presence of the hot pixels results in an overall darker image. (b): Hot pixel removed. (c): Conventional Gaussian noise reduction degrades image details. (d): With our proposed degradation model, the image still maintains faithful details.}
    \label{fig:ablation}
\end{figure}
We analyze the effectiveness of the method mentioned in \cref{sec:degradation_model} through an ablation experiment, shown in \cref{fig:ablation}. 
Before temporal pre-processing, there are several hot pixels in the original image, and these extremely bright white dots pull down the overall brightness of the picture during the grayscale normalization process, which also degrade the grayscale resolution.
Based on the pre-calibrated hot pixel table, we perform temporal pre-processing to remove this fixed pattern noise. In \cref{fig:ablation}b, the overall brightness of the picture is acceptable but presents a lot of random noise. With traditional Gaussian noise reduction method~\cite{liang2021swinir}, (c) exhibits less noise but it also makes some details, such as text in the background, blurry. In (d), by applying a degradation model for both temporal domain and spatial domain, we not only remove a large amount of random noise but also retain most of the details of the texture in the image.


\section{Conclusion}
\label{sec:conclusion}
In this work, we have successfully achieved high-quality event-to-image conversion in static scenes using a stationary event camera. We design a unique event acquisition device, the Adjustable Transmittance Dynamic Vision Sensor (\ourhardware), to capture positive events generated during the gradual increase of transmittance, forming a \timeimage. Subsequently, we map the \timeimage to an adaptive dynamic range grayscale image using the proposed \shortname. We analyze the degradation factors for reconstruction and designed a temporal and spatial degradation model for training.
Experimental results on collected \ourdataset demonstrate the excellent photographic performance of \shortname in high dynamic range and low-light scenarios, along with enhanced performance in computer vision tasks.

\section*{Acknowledgements}
This work was supported in part by the National Key R\&D Program of China, under Grant No.2022YFF0705500, in part by the National Natural Science Foundation of China under Grant No.12174341. 

%
%
\bibliographystyle{splncs04}
\bibliography{main}
\end{document}